\documentclass[pdflatex,sn-mathphys-num]{sn-jnl}

\usepackage{graphicx}%
\usepackage{multirow}%
\usepackage{amsmath,amssymb,amsfonts}%
\usepackage{amsthm}%
\usepackage{mathrsfs}%
\usepackage[title]{appendix}%
\usepackage{textcomp}%
\usepackage{manyfoot}%
\usepackage{booktabs}%
\usepackage{algorithm}%
\usepackage{algorithmicx}%
\usepackage{algpseudocode}%
\usepackage{listings}%
\usepackage{makecell} 
\usepackage[table]{xcolor}%

\theoremstyle{thmstyleone}%
\theoremstyle{thmstyletwo}%

\theoremstyle{thmstylethree}%

\begin{document}

\title[Article Title]{SurgViVQA: Temporally-Grounded Video Question Answering for Surgical Scene Understanding}


\author[1]{\fnm{Mauro Orazio} \sur{Drago}}\email{mauroorazio.drago@mail.polimi.it}
\equalcont{These authors contributed equally to this work.}

\author[1]{\fnm{Luca} \sur{Carlini}}\email{luca.carlini@polimi.it}
\equalcont{These authors contributed equally to this work.}

\author[1]{\fnm{Pelinsu} \sur{Celebi Balyemez}}

\author[1]{\fnm{Dennis} \sur{Pierantozzi}}

\author[1]{\fnm{Chiara} \sur{Lena}}

\author[2]{\fnm{Cesare} \sur{Hassan}}

\author[3]{\fnm{Danail} \sur{Stoyanov}}

\author[1]{\fnm{Elena} \sur{De Momi}}

\author[3]{\fnm{Sophia} \sur{Bano}}

\author[3,4]{\fnm{Mobarak I.} \sur{Hoque}} \email{mobarak.hoque@manchester.ac.uk}

\affil[1]{\orgdiv{Dipartimento di Elettronica, Informazione e Bioingegneria (DEIB)}, \orgname{Politecnico di Milano}, \country{Italy}}

\affil[2]{\orgdiv{IRCCS Humanitas Research Hospital},  
\country{Italy}}

\affil[3]{\orgdiv{UCL Hawkes Institute and Department of Computer Science}, 
\orgname{University College London}, 
\orgaddress{\country{UK}}}

\affil[4]{\orgdiv{Division of Informatics, Imaging and Data Science}, \orgname{University of Manchester}, \orgaddress{\country{UK}}}

\abstract{

\textbf{Purpose}: Video Question Answering (VideoQA) in the surgical domain aims to enhance intraoperative understanding by enabling AI models to reason over temporally coherent events rather than isolated frames. Current approaches are limited to static image features, and available datasets often lack temporal annotations, ignoring the dynamics critical for accurate procedural interpretation.

\textbf{Methods:} 
We propose SurgViVQA, a surgical VideoQA model that extends visual reasoning from static images to dynamic surgical scenes. It uses a Masked Video–Text Encoder to fuse video and question features, capturing temporal cues like motion and tool–tissue interactions, which a fine-tuned LLM then decodes into coherent answers.
To evaluate its performance, we curate REAL-Colon-VQA, a colonoscopic video dataset including motion questions and diagnostic attributes, including out-of-template questions with rephrased or semantically altered formulations to evaluate model robustness.


\textbf{Results}: Experimental validation on REAL-Colon-VQA and the public EndoVis18-VQA dataset shows that SurgViVQA outperforms existing image-based VQA benchmark models and remains competitive with a fine-tuned video VLM baseline. In particular, SurgViVQA improves over PitVQA by +9\% on REAL-Colon-VQA and +9\% on EndoVis18-VQA in Keyword Accuracy, while achieving the strongest overall lexical and semantic generation performance. A perturbation study on the questions further confirms improved generalizability and robustness to variations in question phrasing. 

\textbf{Conclusion:} SurgViVQA and the REAL-Colon-VQA dataset provide a framework for temporally-aware understanding in surgical VideoQA, enabling AI models to interpret dynamic procedural contexts.

}




\maketitle

\section{Introduction}\label{sec:introduction}

During minimally invasive procedures, surgeons must interpret complex and dynamic visual scenes to guide critical decisions \cite{khan2023current}. Understanding these scenes requires modeling not only the spatial content within individual frames but also temporal dependencies across sequences. Subtle temporal cues, such as instrument motion, unfolding anatomy, lesion progression, or endoscope movement, are critical for assessing surgical progress and preventing complications.

General purpose Vision–Language Models (VLMs) are a promising framework for surgical understanding, combining visual perception with textual reasoning, despite limited clinical knowledge \cite{Carlinigutjnl-2025-335091, massimi2025llmparis}. However, trained on natural images, they often miss subtle endoscopic cues (mucosal texture, illumination artifacts, tool–tissue interactions).

However, current surgical Visual Question Answering (VQA) systems remain limited. Foundational approaches, such as SurgicalGPT \cite{Seenivasan2023SurgicalGPT}, extend image–text transformers by integrating visual–text embeddings into a GPT-2 \cite{radford2019gpt2} decoder for autoregressive generation. Similarly, PitVQA \cite{pitvqa} combines a ViT image encoder with a cross-attention text encoder and GPT-2 decoder, grounding visual features into textual tokens. Despite these advances, both models operate on isolated frames, formulate answers as fixed classes or frame-level predictions, and fail to model cause–effect relationships, temporal dependencies, or dynamic aspects of surgical procedures.

The challenge is compounded by the datasets themselves. Most surgical VQA datasets lack temporal annotations and evaluate models in an in-template manner, where test questions closely mirror training examples \cite{surgicalVQA, pitvqa, kvasir-vqa}. This encourages text-biased strategies, limiting the ability to assess model performance under rephrased or semantically varied questions, which are crucial for real-world clinical applications. Consequently, models may generate linguistically plausible answers that fail to convey the key clinical information.



While full fine-tuning is often impractical, parameter-efficient fine-tuning methods such as Low-Rank Adaptation (LoRA) \cite{lora} provide a feasible way to adapt large video VLMs to the surgical domain. However, it remains unclear whether such adaptation alone is sufficient, motivating comparison with a dedicated temporally-aware surgical VideoQA architecture.

In this paper, we overcome these challenges through a temporally-aware VideoQA architecture and a large-scale colonoscopic dataset designed to capture clinically meaningful temporal dynamics.
Our key contributions are as follows:

\begin{itemize}
\item[--] We present \textbf{SurgViVQA}, a modular surgical VideoQA framework that processes short 8-frame clips and enables a controlled study of how different temporal visual encoders affect performance. SurgViVQA fuses tube-masked video tokens with question tokens into unified spatiotemporal representations, which are decoded by a lightweight language decoder (e.g., GPT-2 \cite{radford2019gpt2}, Qwen \cite{qwen}) to generate open-ended answers.

\item[--] We introduce REAL-Colon-VQA, a colonoscopic video dataset annotated with temporally coherent Video Question and Answer (Q\&A) pairs that capture clinically relevant events such as tool usage, lesion motion, and endoscope navigation. The dataset enables both in-template and out-of-template evaluations, including question perturbations designed to test robustness to linguistic variability. This is to our knowledge the first colonoscopy related Video Q\&A dataset publicly available.

\item[--] We extensively evaluate our framework on REAL-Colon-VQA and EndoVis18-VQA against image-based baselines, zero-shot VLMs, and a fine-tuned video VLM baseline, showing that SurgViVQA provides strong and balanced performance in temporal reasoning, semantic robustness, and keyword-level accuracy.

\end{itemize}

\section{Methodology}\label{sec:methodology}

\subsection{Proposed Method: SurgViVQA}

\begin{figure}[h]
    \centering
    \includegraphics[width=\linewidth]{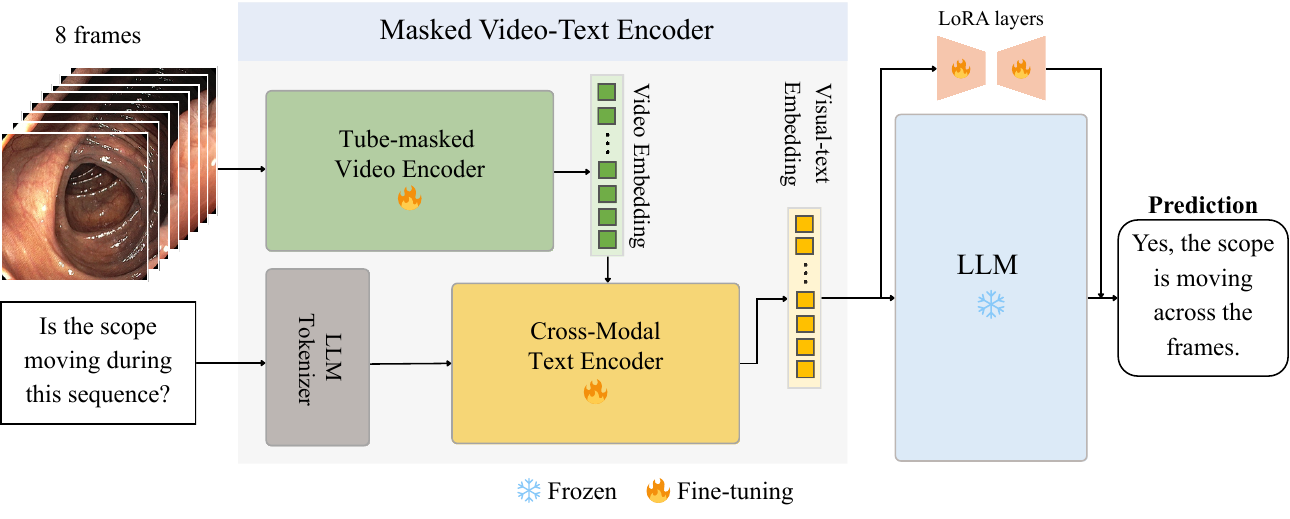}
    \caption{\textbf{SurgViVQA}: The architecture integrates a Masked Video–Text Encoder with a LoRA-adapted LLM to generate context-grounded answers from surgical videos.}
    \label{fig:model}
\end{figure}

As in Fig.~\ref{fig:model}, SurgViVQA processes surgical video clips and questions through a Masked Video–Text Encoder, producing temporally-aware multimodal embeddings that are decoded by a LoRA-adapted LLM to generate context-grounded answers. 

\subsubsection*{Masked-Video Embeddings}

Unlike static images, videos exhibit temporal continuity, meaning that adjacent frames often contain slowly evolving semantics such as instrument motion, camera movement, or tissue deformation. This results in strong temporal redundancy and correlation, where visual content remains largely unchanged between consecutive frames.
Masked visual modeling has been proposed to learn effective visual representations by reconstructing hidden parts of the input \cite{videomae}. In this work, we repurpose VideoMAE from clip-level classification to token-level temporal encoding for surgical VideoQA, conditioning answer generation on temporally consistent visual evidence.
In videos, a clip of $T$ frames of size $H \times W$ can be represented as a sequence of spatiotemporal \textbf{cubes}, where each cube covers a small spatial region over a short temporal span. With cubes of size $t \times h \times w$, the clip forms a token grid of $\frac{T}{t} \times \frac{H}{h} \times \frac{W}{w}$, each represented by a $D$-dimensional feature vector.
To encourage temporal understanding during training, high-ratio masking is applied in \textbf{tubes}, where all cubes at the same spatial location across the temporal dimension are simultaneously hidden. This design forces the model to reconstruct missing content without relying on low-level temporal continuity, instead leveraging high-level semantic reasoning, capturing tool interactions, and anatomical dynamics across time.

\subsubsection*{Architecture}
\noindent \textit{Masked Video-Text Encoder}: To capture temporal dynamics in surgical procedures, Masked Video–Text Encoder consists of a video encoder and a text encoder with a built-in cross-attention mechanism. It integrates tube-masked video embeddings with question representations, allowing each question token to attend to the full temporal sequence of visual features. This produces unified embeddings that capture tool interactions, and anatomical dynamics, suitable for surgical VideoQA.

\noindent \textit{LLM with LoRA Adaptation}: The fused video–text embeddings are passed into a causal, frozen LLM (e.g., GPT-2 \cite{radford2019gpt2}, Qwen \cite{qwen}) enhanced with Low-Rank Adaptation (LoRA) \cite{lora} for efficient fine-tuning. The decoder is built from stacked transformer blocks with multi-head self-attention, layer normalization, and feed-forward layers. Self-attention allows each token to incorporate information from previous tokens, capturing sequential dependencies, while multi-head attention enables the model to attend to different aspects of the input. This module generates open-ended, free-form responses autoregressively, producing coherent and detailed answers grounded in the video–text context.

\subsection{Proposed Dataset: REAL-Colon-VQA}\label{subsec:realcolonvqa}


We introduce REAL-Colon-VQA with the explicit goal of providing a colonoscopy-specific benchmark for temporally grounded surgical VideoQA, by extending the REAL-Colon dataset \cite{biffi2024realcolon}. To support evaluation of clinically relevant endoscopic dynamics, each procedure was reviewed frame-by-frame to annotate endoscope motion (advancing, withdrawing, exiting), tool usage (catheter, snare, forceps), visibility, occlusions, flushing, and illumination mode (white light vs.\ narrow band imaging). Lesion attributes (location, size, histopathology) are inherited from the clinically verified REAL-Colon labels \cite{biffi2024realcolon}.

The frame-level motion-, occlusion- and presence-related labels were produced by two non-clinical annotators with prior experience in endoscopic video analysis, under the supervision of an expert endoscopist, with a third annotator resolving disagreements.
From these annotations we automatically generate QA pairs targeting spatial and temporal reasoning. Questions are instantiated from clinically validated templates and paraphrased into semantically equivalent variants to assess linguistic robustness. Each question has two answers: a short factual answer (e.g.\ ``yes'', ``catheter'', ``advancing'') and a longer descriptive answer providing context.
Each QA pair is linked to an 8-frame video clip. Clips are constructed by sampling one frame every 4 frames from the original 30 fps stream (stride $=4$), resulting in a temporal span of $\approx 0.93$,s. This design intentionally targets short-horizon endoscopic dynamics that are observable within $\sim$1,s, such as scope advancement/withdrawal, brief lesion drift on screen, transient flushing, illumination switches, and short occlusions, while keeping training and inference computationally feasible. A question is assigned to a clip only when the corresponding clinical/visual condition is sustained for the majority of frames in that clip, ensuring that supervision reflects temporal consistency across the sequence rather than a single-frame snapshot.

REAL-Colon-VQA comprises 5{,}200 clip--QA instances and spans 20 question categories across six domains: Instrument presence and type (10.58\%), Sizing (2.88\%), Diagnosis (histopathology; 2.88\%), Positions (anatomical site and on-screen location; 5.77\%), Operation Notes (illumination, visibility, flushing, occlusion; 46.15\%), and Movement (scope movement and lesion motion; 31.73\%). This extends prior surgical VQA datasets such as EndoVis18-VQA \cite{surgicalVQA}, PitVQA \cite{pitvqa}, Kvasir-VQA \cite{kvasir-vqa} and SSG-VQA \cite{ssgvqa} by explicitly modelling temporal events, which are not captured in those datasets (Table~\ref{tab:vqa_comparison_transposed}). Unlike existing surgical VQA benchmarks, operating on isolated frames or frame-level classification, REAL-Colon-VQA links motion cues, visibility conditions and diagnostic attributes to short video clips, providing (to our knowledge) the first colonoscopic VideoQA dataset that supports temporal and diagnostic reasoning in real endoscopic footage.


\begin{figure}
    \centering
    \includegraphics[width=\linewidth]{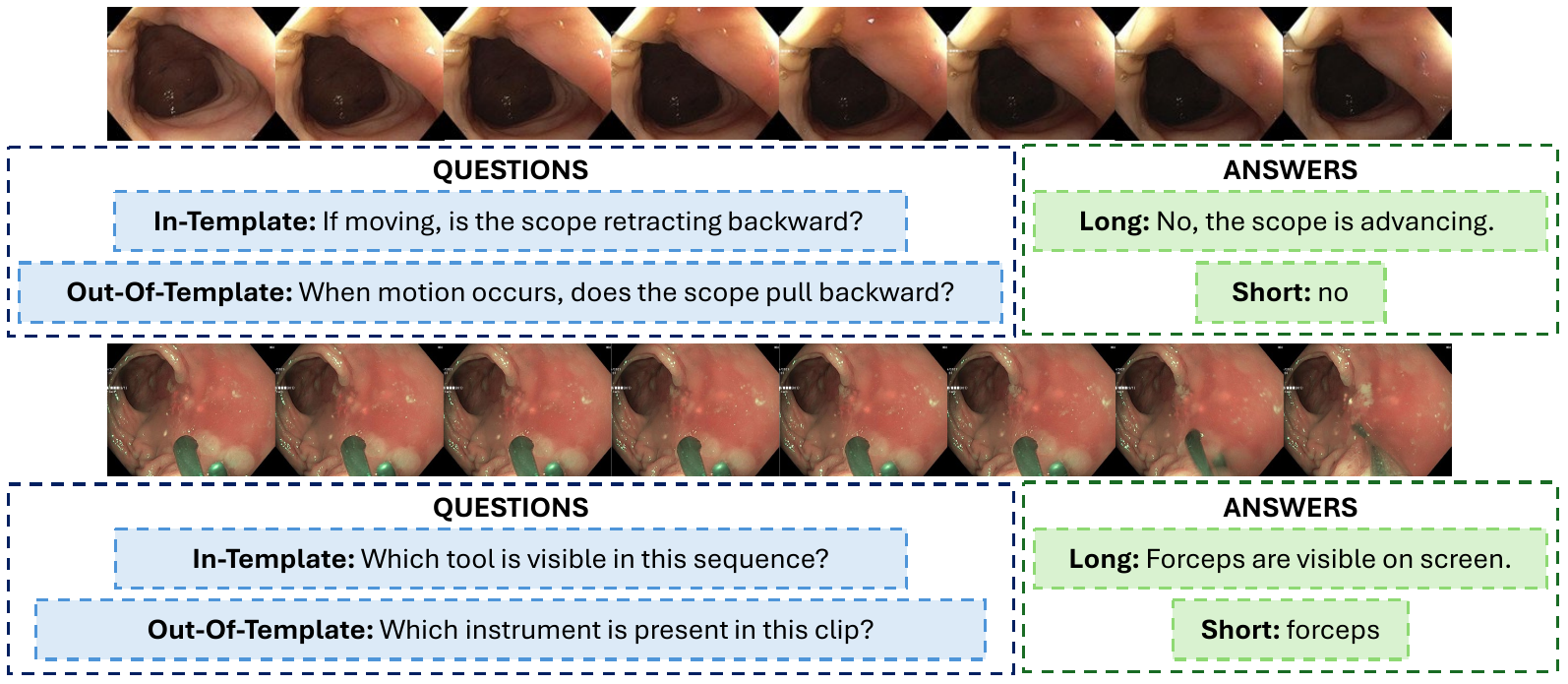}
    \caption{Example from the REAL-Colon-VQA dataset.  }
    \label{fig:datasesamples}
\end{figure}

\begin{table*}[htbp]
\centering
\caption{Comparison of surgical VQA datasets. REAL-Colon-VQA is the only video-based dataset with out-of-template questions, and short/long answers.}
\label{tab:vqa_comparison_transposed}
\resizebox{\linewidth}{!}{%
\begin{tabular}{lccccc}
\toprule
\textbf{Category} 
& \textbf{EndoVis18-VQA \cite{surgicalVQA, qa_snne}} 
& \textbf{PitVQA \cite{pitvqa}} 
& \textbf{Kvasir-VQA \cite{kvasir-vqa}}  
& \textbf{SSG-VQA \cite{ssgvqa}} 
& \textbf{REAL-Colon-VQA} \\
\midrule
\textbf{Procedure} 
& Nephrectomy 
& Transsphenoidal 
& Colon/Gastroscopy 
& Cholecystectomy 
& Colonoscopy \\
\textbf{Video-based} 
& \textcolor{red}{$\times$} 
& \textcolor{red}{$\times$} 
& \textcolor{red}{$\times$} 
& \textcolor{red}{$\times$} 
& \textcolor{green!60!black}{$\checkmark$}  \\  
\textbf{Out-of-template} 
& \textcolor{green!60!black}{$\checkmark$}  
& \textcolor{red}{$\times$} 
& \textcolor{red}{$\times$} 
& \textcolor{red}{$\times$} 
& \textcolor{green!60!black}{$\checkmark$}  \\
\textbf{Short-answer} 
& \textcolor{red}{$\times$} 
& \textcolor{red}{$\times$} 
& \textcolor{red}{$\times$} 
& \textcolor{red}{$\times$} 
& \textcolor{green!60!black}{$\checkmark$}  \\  
\midrule
\textbf{\#Steps} 
& 0 
& 19 
& 1 
& 0 
& 0 \\
\textbf{\#Instruments} 
& 6 
& 18 
& 1 
& 6 
& 3 \\
\textbf{\#Positions} 
& 4 
& 5 
& 3 
& 3 
& 2 \\
\textbf{\#Sizing} 
& 0 
& 0 
& 1 
& 0 
& 1 \\
\textbf{\#Diagnosis} 
& 0 
& 0 
& 2 
& 0 
& 1 \\
\textbf{\#Operation Notes} 
& 13 
& 14 
& 7 
& 10 
& 5 \\
\textbf{\#Movement} 
& 0 
& 0 
& 0 
& 0 
& 5 \\
\bottomrule
\end{tabular}
}
\end{table*}

\section{Experiments and Results}\label{sec:experiments}
\subsection{Dataset}
In addition to our released REAL-Colon-VQA dataset (Section \ref{subsec:realcolonvqa}), split into 4{,}450 training instances and 750 testing instances, we conduct our experiments on the publicly available EndoVis18-VQA dataset \cite{surgicalVQA,qa_snne}, comprising 13,790 QA pairs extracted from 2,086 surgical scenes across 14 nephrectomy videos. The questions address anatomical structures, tool–tissue interactions, and instrument locations, and the answers belong to 18 discrete classes (1 kidney, 13 tool–tissue interactions, and 4 instrument locations). We follow the original official split, consisting of 1,560 frames and 10,574 QAs for training, and 447 frames with 3,216 QAs for validation.
For the image-level setting, we retain the dataset in its original form, while,  for the video-level setting, we construct 8-frame clips and retain only samples for which all 8 consecutive frames are available and aligned with the final frame’s QA annotation. 
To assess robustness beyond template-based formulations, we perform out-of-template evaluation using a perturbed version of the original questions as in \cite{qa_snne}.

\subsection{Implementation Details}

Our SurgViVQA architecture is implemented in PyTorch using HuggingFace repositories, with a Masked Video–Text Encoder built from pre-trained Video Masked Autoencoders (VideoMAE) \cite{videomae} video encoder and the Bootstrapping Language-Image Pre-training (BLIP) cross-attention text encoder \cite{blip} to fuse video and question representations. To adapt VideoMAE to the 8-frame setting, the patch-embedding and final LayerNorm layers are kept unfrozen, while the rest of the encoder remains frozen, as pre-training was done with 16 frames. The LLM is fine-tuned on the task using LoRA. For GPT-2, the low-rank adapters are applied to the attention (c\_attn) and projection (c\_proj) layers, while for Qwen the adapters target the query, key, value, and output projection layers (q\_proj, k\_proj, v\_proj, o\_proj).
The model is trained for 60 epochs using the Adam optimizer with a learning rate of $2\times10^{-7}$, and optimized with Binary Cross-Entropy (BCE) loss. 

For fairness, all comparative SOTA baselines, such as SurgicalGPT and PitVQA, designed for single-frame inputs, are retrained using their official repositories. We additionally evaluate recent general-purpose VLMs (InternVL3-1B \cite{internvl3} and Qwen2.5-VL-3B \cite{bai2025qwen25vltechnicalreport}) in a zero-shot setting using an extensive prompt describing the surgical environment. To provide a task-adapted video baseline, we also fine-tune InternVL3 \cite{internvl3} with LoRA on each target dataset using the same training split, and report it as a fine-tuned video VLM baseline. Experiments are conducted on NVIDIA H100.

\subsection{Results}

To comprehensively evaluate SurgViVQA, we employ a suite of automatic metrics capturing both lexical precision and semantic fidelity. BLEU-4 (BLE-4) measures 4-gram overlap between generated and reference answers, reflecting linguistic accuracy, while ROUGE-L (ROU-L) and METEOR (MET) assess semantic coherence with human responses.
Given the clinical context, we place particular emphasis on Keyword Accuracy (K-ACC), which evaluates whether the model explicitly generates the correct surgical or anatomical term within its response, providing a direct measure of factual grounding and clinical relevance.
Table~\ref{tab:results_table} reports quantitative comparisons on EndoVis18-VQA and REAL-Colon-VQA, including both in-template and out-of-template evaluations for robustness and generalization.

\begin{table}[t]
\centering
\caption{Performance comparison of baseline models, zero-shot methods, PEFT fine-tuned (FT) video models, and our SurgViVQA model across datasets, reporting in-template and out-of-template metrics. All metrics are in percentages (\%). Best results in \textbf{bold}; second-best \underline{underlined}.}
\label{tab:results_table}
\renewcommand{\arraystretch}{1}
\setlength{\tabcolsep}{2.2pt}
\begin{tabular}{@{}llllcccccccc@{}}
\toprule
\multicolumn{12}{c}{\textbf{REAL-Colon-VQA}} \\
\midrule
 & Model & LLM &
& \multicolumn{4}{c}{In-template}
& \multicolumn{4}{c}{Out-of-template} \\
\cmidrule(lr){5-8} \cmidrule(lr){9-12}
& & & & BLE-4 & ROU-L & MET & K-ACC & BLE-4 & ROU-L & MET & K-ACC \\
\midrule
\multirow{2}{*}{\rotatebox{90}{Base}}
& SurgicalGPT \cite{Seenivasan2023SurgicalGPT} & GPT-2 & Image & 14.93 & 47.85 & 52.36 & 33.33 & 12.35 & 42.87 & 50.49 & 46.67 \\
& PitVQA \cite{pitvqa} & GPT-2 & Image & 64.55 & \underline{78.48} & \underline{79.99} & 54.13 & 23.63 & 50.03 & 53.22 & 42.93 \\
\midrule
\multirow{4}{*}{\rotatebox{90}{Zero-Shot}}
& Qwen2.5 \cite{bai2025qwen25vltechnicalreport} & - & Image & 4.53 & 45.70 & 46.83 & 49.33 & 0.50 & 38.80 & 40.77 & 50.27 \\
& MedGemma \cite{medgemma} & - & Image & 18.09 & 36.92 & 31.86 & 49.47 & 5.31 & 30.96 & 27.80 & 34.13 \\
& InternVL3 \cite{internvl3} & - & Video & 7.13 & 43.90 & 54.31 & \textbf{68.67} & 3.87 & 32.10 & 44.34 & \textbf{53.73} \\
& VideoLLaMA3 \cite{videollama3} & - & Video & 2.74 & 33.01 & 42.22 & 17.60 & 1.78 & 26.17 & 35.15 & 21.47 \\
\midrule
\multirow{1}{*}{\rotatebox{90}{FT}}
& InternVL3 + LoRA & - & Video & \underline{64.88} & 75.15 & 76.42 & 59.07 & \textbf{32.97} & \underline{52.37} & \underline{54.37} & \underline{51.77} \\
\midrule
\multirow{2}{*}{\rotatebox{90}{Ours}}
& SurgViVQA & Qwen & Video & 60.09 & 74.36 & 75.32 & 59.73 & 16.84 & 37.69 & 35.26 & 38.00 \\
& \textbf{SurgViVQA} & GPT-2 & Video & \textbf{71.98} & \textbf{82.85} & \textbf{84.11} & \underline{62.80} & \underline{31.19} & \textbf{53.62} & \textbf{54.89} & 47.73 \\
\midrule
\multicolumn{12}{c}{\textbf{EndoVis18-VQA}} \\
\midrule
& Model & LLM &
& \multicolumn{4}{c}{In-template}
& \multicolumn{4}{c}{Out-of-template \cite{qa_snne}} \\
\cmidrule(lr){5-8} \cmidrule(lr){9-12}
& & & & BLE-4 & ROU-L & MET & K-ACC & BLE-4 & ROU-L & MET & K-ACC \\
\midrule
\multirow{2}{*}{\rotatebox{90}{Base}}
& SurgicalGPT \cite{Seenivasan2023SurgicalGPT} & GPT-2 & Image & 29.55 & 58.60 & 57.99 & 4.00 & 2.52 & 43.97 & 44.99 & 11.85 \\
& PitVQA \cite{pitvqa} & GPT-2 & Image & \underline{81.73} & 86.18 & 83.28 & 40.35 & \underline{17.57} & 46.87 & 45.49 & 9.73 \\
\midrule
\multirow{4}{*}{\rotatebox{90}{Zero-Shot}}
& Qwen2.5 \cite{bai2025qwen25vltechnicalreport} & - & Image & 19.04 & 57.60 & 68.60 & \underline{51.22} & 11.92 & \textbf{52.77} & \textbf{63.72} & \underline{46.90} \\
& MedGemma \cite{medgemma} & - & Image & 8.23 & 44.32 & 64.19 & 37.90 & 7.74 & 40.74 & \underline{61.59} & 37.86 \\
& InternVL3 \cite{internvl3} & - & Video & 5.62 & 51.50 & 55.33 & 32.21 & 0.57 & 38.73 & 38.73 & 21.99 \\
& VideoLLaMA3 \cite{videollama3} & - & Video & 3.28 & 52.01 & 54.66 & 40.90 & 0.57 & 38.73 & 39.51 & 35.89 \\
\midrule
\multirow{1}{*}{\rotatebox{90}{FT}}
& InternVL3 + LoRA & - & Video & 33.99 & 72.62 & 83.67 & \textbf{70.31} & 6.27 & 28.04 & 37.08 & \textbf{56.42} \\
\midrule
\multirow{2}{*}{\rotatebox{90}{Ours}}
& SurgViVQA & Qwen & Video & 79.05 & \underline{87.98} & \underline{84.70} & 46.30 & 0.60 & 12.33 & 8.10 & 0.99 \\
& \textbf{SurgViVQA} & GPT-2 & Video & \textbf{84.94} & \textbf{89.65} & \textbf{86.08} & 48.27 & \textbf{24.84} & \underline{50.86} & 50.13 & 9.23 \\
\bottomrule
\end{tabular}
\end{table}

\begin{figure}[h]
    \centering
        \includegraphics[width=\linewidth]{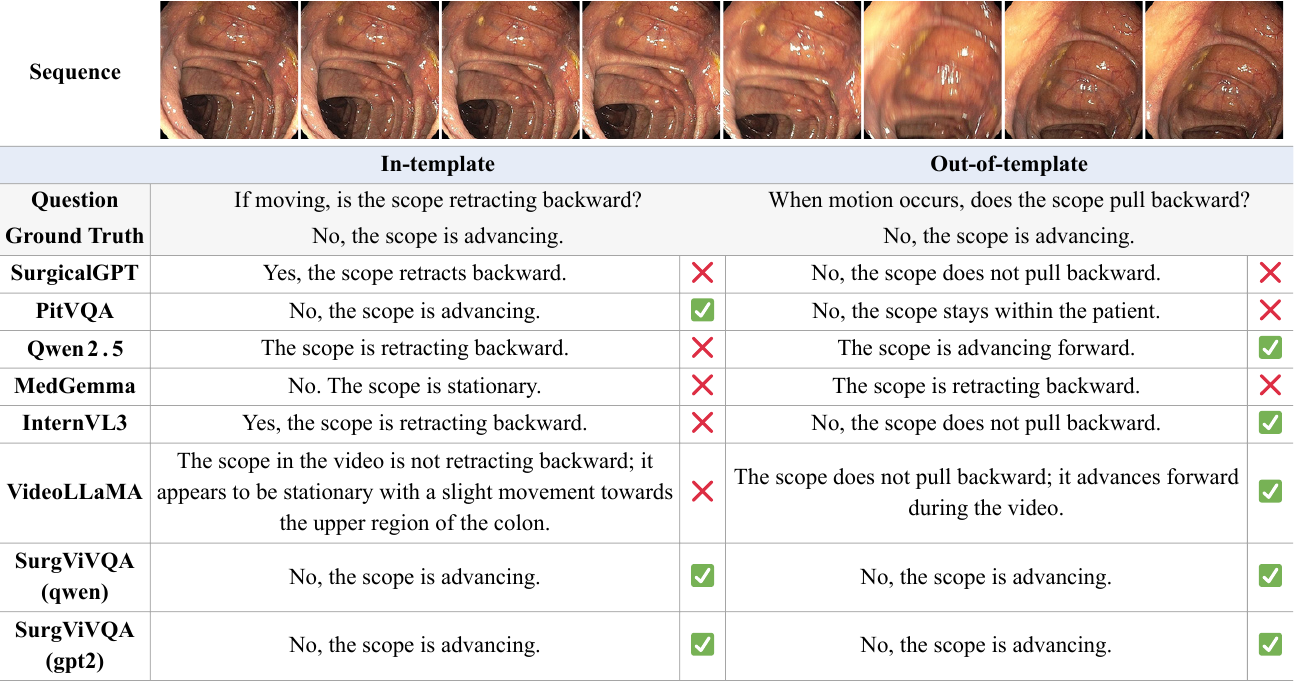}
    \caption{Qualitative comparison of predictions from all models on in-template and out-of-template example from the REAL-Colon-VQA dataset}
    \label{fig:qualitative_results}
\end{figure}

\begin{table}[htbp]
\centering
\caption{In-template performance on REAL-Colon-VQA, reporting Keyword Accuracy (K-ACC, \%) overall and stratified by motion-aware and static questions.}
\label{tab:static_vs_dynamic}
\setlength{\tabcolsep}{13pt}
\begin{tabular}{@{}l c c c@{}}
\toprule
Model 
& Overall K-ACC 
& Motion-aware K-ACC 
& Static K-ACC \\
\midrule
PitVQA \cite{pitvqa} (Image)            & 54.13 & 42.00 & 58.55 \\
InternVL3 \cite{internvl3} (Video)        & 68.67 & 82.00 & 63.82 \\
SurgViVQA (Qwen)         & 59.73 & 83.00 & 51.27 \\
SurgViVQA (GPT-2)        & 62.80 & 62.50 & 62.91 \\
\bottomrule
\end{tabular}
\end{table}

For in-template performance, SurgViVQA with GPT-2 remains the strongest overall generative model, achieving the best BLEU-4, ROUGE-L, and METEOR scores on both EndoVis18-VQA and REAL-Colon-VQA, with Keyword Accuracy of 48.3\% and 62.80\%, respectively. Compared with PitVQA, this corresponds to improvements of roughly 9\% on EndoVis18-VQA and 9\% on REAL-Colon-VQA in K-ACC, showing that factual grounding gains do not come at the expense of linguistic fluency.
The fine-tuned video baseline, InternVL3 + LoRA, further strengthens the comparison. On REAL-Colon-VQA, SurgViVQA still achieves the best overall in-template performance, while on EndoVis18-VQA InternVL3 + LoRA attains the highest K-ACC, indicating that PEFT adaptation of a general-purpose video VLM can substantially improve task performance, although its effect differs across datasets and metrics. However, SurgViVQA maintains markedly stronger lexical and semantic generation quality, suggesting that a dedicated surgical VideoQA architecture provides a better balance between factual grounding and open-ended answer generation.
For out-of-template evaluations, all models show a drop in performance, reflecting the challenge of generalizing to unseen question types. The fine-tuned InternVL3 baseline improves substantially over its zero-shot counterpart, particularly on REAL-Colon-VQA, confirming the value of task adaptation for video VLMs. Nevertheless, SurgViVQA with GPT-2 remains competitive in out-of-template language generation, especially on EndoVis18-VQA where it achieves the highest BLEU-4, while zero-shot Qwen2.5 attains higher ROUGE-L and METEOR. Fig.~\ref{fig:qualitative_results} presents a qualitative comparison of the predictions produced by all evaluated models on both in-template and out-of-template samples from the REAL-Colon-VQA dataset.

To analyze failure modes, we partition REAL-Colon-VQA into motion-aware questions that require temporal evidence across frames (motion-aware) and static questions that can be answered from a single frame. Table~\ref{tab:static_vs_dynamic} reports in-template K-ACC for the best image baseline (PitVQA), the best video baseline (InternVL3), and both our variants (SurgViVQA--Qwen/GPT-2). Video models dominate on motion-aware queries (InternVL3 82.0\%, SurgViVQA--Qwen 83.0\% vs.\ PitVQA 42.0\%), while static questions narrow the gap and highlight that Qwen is weaker on static cues (PitVQA 58.6\% vs.\ SurgViVQA--Qwen 51.3\%), consistent with its poorer EndoVis performance.

\subsection{Ablation Study}

We evaluate our method on the EndoVis18-VQA dataset through two complementary ablation studies: (1) analysis of different temporal feature extractors as video encoders paired with GPT-2, and (2) evaluation of keyword penalization in the weighted BCE loss to mitigate missing critical keywords in generated answers.

\noindent \textit{(1) Temporal Feature Extractors:}
A core contribution of this work is the systematic analysis of temporal visual encoders for surgical VideoQA. To isolate the role of the Temporal Feature Extractor (Temporal FE), we perform a controlled ablation on the in-template EndoVis18-VQA split, comparing X-CLIP \cite{xclip}, ViViT \cite{vivit}, TimeSformer \cite{timesformer}, VideoMamba \cite{videomamba}, and VideoMAE \cite{videomae}. All variants use the same GPT-2 decoder and identical training protocol, so differences can be attributed to the Temporal FE. Results are reported in Table~\ref{tab:tfe_ablation_study}.
Our experiments show that VideoMAE consistently achieves the best performance across BLEU, ROUGE, and METEOR metrics, highlighting its superior ability to capture spatiotemporal features in surgical video clips. 
VideoMamba provides competitive results, but falls
slightly behind VideoMAE, emphasizing the importance of encoder design for effective video representation. Overall, the results demonstrate that selecting a high-capacity video encoder is crucial for generative surgical VQA performance.

\begin{table}[t]
\centering
\caption{Performance of different video encoders paired with GPT-2 on in-template EndoVis18-VQA. All metrics are reported in percentages (\%). Best results are in \textbf{bold}; second-best results are \underline{underlined}.}
\setlength{\tabcolsep}{13pt}
\begin{tabular}{@{}l c c c c c@{}}
\toprule
Temporal FE 
& BLEU-3 & BLEU-4 & ROUGE-2 & ROUGE-L & METEOR \\
\midrule
X\,-CLIP \cite{xclip}       & 83.41 & 78.81 & \underline{85.81} & 88.80 & \textbf{86.79} \\
ViViT \cite{vivit}           & 84.08 & 78.27 & 82.74 & 86.89 & 84.02 \\
TimeSformer \cite{timesformer}     & 84.73 & 81.02 & 82.63 & 86.31 & 83.02 \\
VideoMAE \cite{videomae}       & \textbf{88.44} & \textbf{84.94} & \textbf{89.62} & \textbf{89.66} & \underline{86.08} \\
VideoMamba \cite{videomamba}     & \underline{87.96} & \underline{84.47} & 85.76 & \underline{89.06} & 85.64 \\
\bottomrule
\end{tabular}
\label{tab:tfe_ablation_study}
\end{table}
\noindent \textit{(2) Keyword Penalization:}
To address the issue of missing crucial keywords in generated answers (e.g., predicting "lung" instead of "kidney"), we introduce a penalization term in the weighted BCE loss. Let $y_i \in {0,1}$ denote the reference label for token $i$, $\hat{y}_i$ the predicted probability, and $\mathcal{K}$ the set of keywords in the reference. The modified loss is defined as:
\begin{equation}
\mathcal{L}_{\text{WBCE}} = \frac{1}{N} \sum_{i=1}^{N} w_i \, \text{BCE}(y_i, \hat{y}_i),
\qquad
w_i =
\begin{cases}
\lambda & \text{if token $i$ is a keyword} \\
1 & \text{otherwise}
\end{cases}
\end{equation}
where $N$ is the total number of tokens in the answer and $\lambda$ is the keyword weighting factor, which increases the penalty for mispredicting keywords, encouraging the model to focus on correctly predicting critical keywords.

\begin{table}[t]
\centering
\caption{Performance of SurgViVQA with GPT-2 as LLM and different keyword penalization $\lambda$ in the weighted BCE loss on EndoVis18-VQA dataset. Metrics are reported in percentages (\%). Best results in \textbf{bold}; second-best results are \underline{underlined}.}
\label{tab:keyword_ablation}
\renewcommand{\arraystretch}{1}
\setlength{\tabcolsep}{6pt} 
\begin{tabular}{@{}l c c c c c c c c c@{}}
\toprule
\multicolumn{2}{c}{} 
& \multicolumn{4}{c}{In-template} 
& \multicolumn{4}{c}{Out-of-template} \\
\cmidrule(lr){3-6} \cmidrule(lr){7-10}
Model & $\lambda$ 
& BLE-4 & ROU-L & MET & K-ACC & BLE-4 & ROU-L & MET & K-ACC \\ 
\midrule
\multirow{6}{*}{\centering \makecell[l]{SurgViVQA \\ (GPT-2)}} 
& 1  & \textbf{84.94 }& 89.65 & 86.08 & 48.27 & 24.85 & 50.86 & 50.13 & 9.23 \\
& 2  & \underline{84.71} & 89.75 & 86.17 & \underline{48.61} & 25.14 & 50.92 & 50.13 & 9.12 \\
& 5  & 84.53 & 90.71 & 87.83 & 48.42 & \underline{25.68} & \underline{51.08} & \underline{50.14} & 8.81 \\
& 10 & 83.92 & \textbf{91.87} & \textbf{89.78} & \textbf{49.03} & \textbf{27.91} & \textbf{52.09} & \textbf{50.76} & \underline{11.47} \\
& 25 & 80.31 & 90.81 & \underline{88.66} & 46.45 & 25.14 & 50.92 & 50.13 & 9.12 \\
& 50 & 80.16 & \underline{90.93} & 88.65 & 46.83 & 12.38 & 43.80 & 46.70 & \textbf{12.08} \\
\bottomrule
\end{tabular}
\end{table}

We experiment with different values of $\lambda$, specifically $[1,2,5,10,25,50]$, and compare their effect on keyword accuracy and overall language generation performance using our SurgViVQA with GPT-2 as LLM on EndoVis18-VQA dataset. The setting $\lambda = 1$ corresponds to the standard BCE loss formulation, i.e., without the inclusion of the penalization term.
As shown in Table~\ref{tab:keyword_ablation}, incorporating keyword penalization demonstrates promising potential. Moderate penalization ($\lambda=10$) yields the best trade-off between language fluency and keyword accuracy, while excessively large values degrade performance, confirming that overemphasizing keywords harms overall generation quality. However, with proper tuning of $\lambda$, this technique could be further leveraged to improve both keyword accuracy and overall generation quality.

\section{Discussions and Conclusions}
In this study, we introduced SurgViVQA, a temporally-aware VideoQA model for dynamic surgical procedures, along with REAL-Colon-VQA, a colonoscopic video dataset annotated with temporally coherent Q\&A pairs. SurgViVQA integrates a Masked Video–Text Encoder to combine short video clips with question features, capturing subtle motion and tool–tissue interactions, which are then decoded by a fine-tuned LLM into contextually coherent answers. Ablation studies indicate that tube-masked video encoders most effectively capture spatiotemporal dynamics, while keyword penalization in the loss function further improves the accurate generation of clinically critical terms, together contributing to the enhanced performance of SurgViVQA.
Evaluations on REAL-Colon-VQA and EndoVis18-VQA demonstrate that SurgViVQA consistently outperforms image-based baselines across lexical, semantic, and clinically grounded metrics, including Keyword Accuracy. The additional comparison with a fine-tuned InternVL3 + LoRA baseline further shows that PEFT adaptation can substantially strengthen general-purpose video VLMs, although its gains are metric- and dataset-dependent, a dedicated domain-specific VideoQA architecture remains advantageous for balanced open-ended answer generation. Notably, out-of-template results indicate that the model generalizes better than image-only baselines, although performance decreases under unseen linguistic formulations, highlighting the potential benefits of more diverse question sampling to further improve robustness.

Performance differences are strongly influenced by the type of questions being asked. Image-only baselines struggle with motion semantics, while video models benefit from short-term temporal context. Across SurgViVQA variants, Qwen-based decoding is weaker on static cues (e.g., subtle tool visibility or illumination), as reflected on REAL-Colon-VQA (Table~\ref{tab:static_vs_dynamic}). On EndoVis18-VQA, all SurgViVQA models are constrained by its static questions with single-frame annotations adapted to 8-frame clips, limiting gains from temporal modeling and likely explaining the observed drop in K-ACC compared to the REAL-Colon-VQA dataset. Overall, the GPT-2 variant shows more balanced behavior across motion-aware and static questions, indicating stronger robustness to frame-local visual cues than the Qwen variant (Table~\ref{tab:static_vs_dynamic}).

The results reported in Table~\ref{tab:static_vs_dynamic} also show that the time-related questions in our dataset are correctly defined, and that the chosen temporal interval of the video clips is appropriate to capture the targeted motion cues.

Overall, these results highlight the critical role of temporal modeling in surgical VideoQA, showing that integrating motion and sequence dynamics improves both factual accuracy and linguistic fluency. 
Future work will focus on enhancing out-of-domain generalization through more diverse question formulations and extending temporal reasoning beyond 8 frames, aiming to provide more reliable, context-aware decision support for surgeons.
Additionally, future work will explore broader fine-tuning strategies for larger video VLMs and alternative PEFT schemes beyond LoRA, together with a dedicated clinical study with endoscopists to assess usability and practical impact in realistic workflow settings.



\section*{Statements and Declarations}

\bmhead{Author Contributions} 
Mauro Orazio Drago and Luca Carlini contributed equally to this work and share first authorship.

\bmhead{Competing interests}
Authors declare they have no competing financial or non-financial interests directly or indirectly related to the work submitted for publication.

\bmhead{Ethics Approval} 
This article does not contain any new studies with human participants or animals performed by any of the authors. All analyses were conducted on previously published or fully anonymized surgical video datasets, including EndoVis18-VQA and the REAL-Colon dataset from which REAL-Colon-VQA was derived.

\bmhead{Informed Consent}
No new data involving human participants were collected for this study. Informed consent was obtained by the original investigators at the time of data acquisition, where applicable; therefore, additional informed consent for this secondary analysis was not required.

\bmhead{Data and code availability}
Dataset and code are available in our \href{https://github.com/madratak/SurgViVQA}{GitHub}.

\bmhead{Funding}
This work was supported by the Multilayered Urban Sustainability Action (MUSA) project (ECS00000037), funded by the European Union – NextGenerationEU under the National Recovery and Resilience Plan (NRRP); the ANTHEM project, funded by the National Plan for NRRP Complementary Investments (CUP: B53C22006700001); the Engineering and Physical Sciences Research Council (EPSRC) [EP/W00805X/1; UKRI145; EP/Y01958X/1]; the Wellcome/EPSRC Centre for Interventional and Surgical Sciences (WEISS) [203145/Z/16/Z]; and the Department for Science, Innovation and Technology (DSIT) and the Royal Academy of Engineering under the Chair in Emerging Technologies programme. For the purpose of open access, the author has applied a CC BY public copyright licence to any Author Accepted Manuscript arising from this submission.

\bibliography{temporary-bibliography}

\end{document}